\pdfoutput=1

\documentclass[11pt]{article}

\usepackage{emnlp2021}

\usepackage{times}
\usepackage{latexsym}

\usepackage{graphicx}
\usepackage{url}
\usepackage{multirow}
\usepackage{booktabs}
\usepackage{verbatim}
\usepackage{amsmath}
\usepackage{subfigure}

\usepackage{amssymb}
\usepackage{pifont}
\newcommand{\cmark}{\ding{51}}%
\newcommand{\xmark}{\ding{55}}%

\usepackage[T1]{fontenc}

\usepackage[utf8]{inputenc}

\usepackage{microtype}

\title{DuRecDial 2.0: A Bilingual Parallel Corpus for Conversational Recommendation}

\author{
	Zeming Liu\textsuperscript{1}\thanks{\quad This work was done at Baidu.}, Haifeng Wang\textsuperscript{2}, Zheng-Yu Niu\textsuperscript{2}, Hua Wu\textsuperscript{2}, Wanxiang Che\textsuperscript{1}\thanks{\quad Corresponding author: Wanxiang Che.}\\
	\textsuperscript{1}Research Center for Social Computing and Information Retrieval,\\
	Harbin Institute of Technology, Harbin, China \\
	\textsuperscript{2}Baidu Inc., Beijing, China \\
	{\tt \{zmliu, car\}@ir.hit.edu.cn} \\
	{\tt \{wanghaifeng, niuzhengyu, wu\_hua\}@baidu.com} \\
}

\begin{document}
\maketitle
\begin{abstract}
In this paper, we provide a bilingual parallel human-to-human recommendation dialog dataset (\textbf{DuRecDial 2.0}) to enable researchers to explore a challenging task of multilingual and cross-lingual conversational recommendation. The difference between \emph{DuRecDial 2.0} and existing conversational recommendation datasets is that the data item (Profile, Goal, Knowledge, Context, Response) in \emph{DuRecDial 2.0} is annotated in two languages, both English and Chinese, while other datasets are built with the setting of a single language.
We collect 8.2k dialogs aligned across English and Chinese languages (16.5k dialogs and 255k utterances in total) that are annotated by crowdsourced workers with strict quality control procedure.
We then build monolingual, multilingual, and cross-lingual conversational recommendation baselines on \emph{DuRecDial 2.0}. 
Experiment results show that the use of additional English data can bring performance improvement for Chinese conversational recommendation, indicating the benefits of \emph{DuRecDial 2.0}. Finally, this dataset provides a challenging testbed for future studies of monolingual, multilingual, and cross-lingual conversational recommendation. \footnote{https://github.com/liuzeming01/DuRecDial.}
\end{abstract}
\section{Introduction}

In recent years, there has been a significant increase in the research topic of conversational recommendation due to the rise of voice-based bots \cite{Kang2019,Li2018,Sun2018,Christakopoulou2016,Warnestal2005}. These works focus on how to provide recommendation service in a more user-friendly manner through dialog-based interactions. They fall into two categories: (1) task-oriented dialog-modeling approaches with requirement of pre-defined user intents and slots \cite{Warnestal2005,Christakopoulou2016,Sun2018}; (2) non-task dialog-modeling approaches that can conduct more free-form interactions for recommendation, without pre-defined user intents and slots \cite{Li2018,Kang2019}. Recently more and more efforts are devoted to the research line of the second category and many datasets have been created, including English dialog datasets \cite{Dodge2016,Li2018,Kang2019,Moon2019,Hayati2020INSPIREDTS} and Chinese dialog datasets \cite{liu-etal-2020-towards,Zhou2020TowardsTC}.

\begin{figure*}[t]
	\centering\includegraphics[ height=3.6in, width=6.3in]{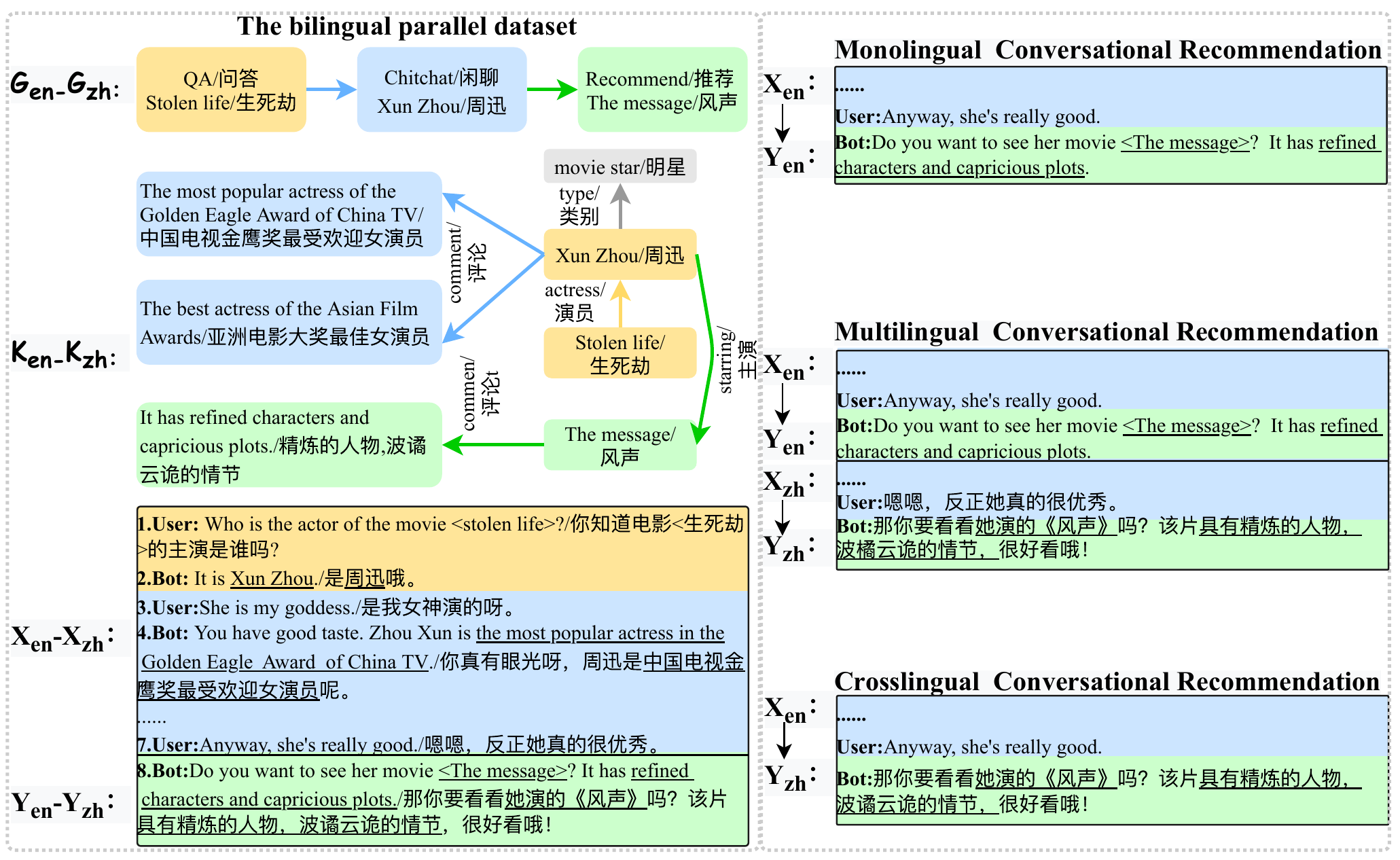}
	\caption{Illustration of DuRecDial 2.0 with the $monolingual$, $multilingual$, and $crosslingual$ conversational recommendation on the dataset. We use different colors to indicate different goals. $\emph{G}$, $\emph{K}$, X, and Y stands for dialog goal, knowledge, context, and response respectively.}\label{fig:example}
\end{figure*}

However, to the best of our knowledge, almost all these datasets are constructed in the setting of a single language, and there is no publicly available multilingual dataset for conversational recommendation. 
Previous work on other NLP tasks have proved that multilingual corpora can bring performance improvement in comparison with monolingual task setting, such as for the tasks of task-oriented dialog \cite{schuster-etal-2019-cross-lingual}, semantic parsing \cite{li-etal-2021-mtop}, QA and reading comprehension \cite{jing-etal-2019-bipar,Lewis2020MLQAEC,artetxe-etal-2020-cross,clark-etal-2020-tydi,Hu2020XTREMEAM,hardalov-etal-2020-exams}, machine translation\cite{johnson-etal-2017-googles}, document classification \cite{Lewis2004,Klementiev2012InducingCD,schwenk-li-2018-corpus}, semantic role labelling \cite{akbik-etal-2015-generating} and NLI \cite{Conneau2018XNLIEC}.
Therefore it is necessary to create multilingual conversational recommendation dataset that might enhance model performance when compared with monolingual training setting, and it could provide a new benchmark dataset for the study of multilingual modeling techniques.

To facilitate the study of this challenge, we present a bilingual parallel recommendation dialog dataset, \textbf{DuRecDial 2.0}, 
for multilingual and cross-lingual conversational recommendation. \emph{DuRecDial 2.0} consists of 8.2K dialogs aligned across two languages, English and Chinese (16.5K dialogs and 255K utterances in total). 
Table \ref{table:datasets} shows the difference between \emph{DuRecDial 2.0} and existing conversational recommendation datasets. We also analyze \emph{DuRecDial 2.0} in-depth and find that it offers more diversified prefixes of utterances and then more flexible language style, as shown in Figure ~\ref{fig:prefixes}(a) and Figure ~\ref{fig:prefixes}(b).

We define five tasks on this dataset. As shown in Figure ~\ref{fig:example} \emph{$Monolingual$}, the first two tasks are English or Chinese monolingual conversational recommendation, where dialog context, knowledge, dialog goal, and response are in the same language. It aims at investigating the performance variation of the same model across two different languages. As shown in Figure ~\ref{fig:example} \emph{$Multilingual$}, there is another task that is called multilingual conversational recommendation. Here we directly mix training instances of the two languages into a single training set and train a single model to handle both English and Chinese conversational recommendation at the same time. As shown in Figure ~\ref{fig:example} \emph{$Crosslingual$}, the last two tasks are cross-lingual conversational recommendation, where model input and output are in different languages, e.g. dialog context is in English (or Chinese) and generated response is in Chinese (or English).

To address these tasks, we build baselines using XNLG \cite{Chi2020CrossLingualNL}\footnote{https://github.com/CZWin32768/XNLG} and mBART \cite{Liu2020MultilingualDP}\footnote{https://github.com/pytorch/fairseq/}. 
We conduct an empirical study of the baselines on \emph{DuRecDial 2.0}, and experiment results indicate that the use of additional English data can bring performance improvement for Chinese conversational recommendation.



In summary, this work makes the following contributions: 
\begin{itemize}
	\item To facilitate the study of multilingual and cross-lingual conversational recommendation, we create a novel dataset \emph{DuRecDial 2.0}, the first publicly available bilingual parallel dataset for conversational recommendation.
	\item We define five tasks, including monolingual, multilingual, and cross-lingual conversational recommendation, based on \emph{DuRecDial 2.0}.
	\item We establish monolingual, multilingual, and cross-lingual conversational recommendation baselines on \emph{DuRecDial 2.0}. The results of automatic evaluation and human evaluation confirm the benefits of this bilingual dataset for Chinese conversational recommendation.
\end{itemize}

\begin{table*}
		\centering
		\small
		\begin{tabular}{ p{4.35cm} r  r r r p{1.7cm} p{2.7cm}} 
			\toprule[1.0pt]
			Datasets  & Language & Parallel & \#Dial. & \#Utt. & Dialog types  &  Domains  \\ 
			\toprule[1.0pt]
			Facebook\_Rec\cite{Dodge2016} & EN& \xmark & 1M & 6M & Rec. & Movie \\  
			REDIAL \cite{Li2018} &  EN& \xmark & 10k& 163k& Rec., chitchat& Movie \\ 
			GoRecDial \cite{Kang2019} &  EN&  \xmark & 9k& 170k& Rec.& Movie \\ 
			OpenDialKG \cite{Moon2019} &  EN&  \xmark & 12k& 143k& Rec.& Movie, book \\ 
			DuRecDial \cite{liu-etal-2020-towards} &  ZH& \xmark & 10.2k& 156k & Rec., chitchat, QA, task & Movie, music, star, food, restaurant, news, weather\\
			TG-ReDial \cite{Zhou2020TowardsTC} & ZH& \xmark & 10k & 129k & Rec. & Movie \\
			INSPIRED \cite{Hayati2020INSPIREDTS} & EN& \xmark & 1k &35k & Rec. &Movie\\
			\hline
			DuRecDial 2.0 (Ours) &  EN-ZH& \cmark & 16.5k& 255k & Rec., chitchat, QA, task & Movie, music, star, food, restaurant, weather\\ \bottomrule[1.0pt]
		\end{tabular}
		\caption{Comparison of \emph{DuRecDial 2.0} with other datasets for conversational recommendation. ``EN'',  ``ZH'', ``Dial.'',  ``Utt.'',  and ``Rec.'' stands for English, Chinese, dialogs, utterances, and recommendation respectively.}
		\label{table:datasets}
	\end{table*}

\section{Related Work}

\textbf{Datasets for Conversational Recommendation} To facilitate the study of conversational recommendation, multiple datasets have been created in previous work, as shown in Table \ref{table:datasets}. The first recommendation dialog dataset is released by \citet{Dodge2016}, which is a synthetic dialog dataset built with the use of the classic MovieLens ratings dataset and natural language templates. \citet{Li2018} creates a human-to-human multi-turn recommendation dialog dataset, which combines the elements of social chitchat and recommendation dialogs. \citet{Kang2019} provides a recommendation dialogue dataset with clear goals, and \citet{Moon2019} collects a parallel Dialog$\leftrightarrow$KG corpus for recommendation. \cite{liu-etal-2020-towards} constructs a human-to-human conversational recommendation dataset contains 4 dialog types and 7 domains, which has clear goals to achieve during each conversation, and user profiles for personalized conversation. \cite{Zhou2020TowardsTC} automatically collects a conversational recommendation dataset, which is built with the use of movie data.   \cite{Hayati2020INSPIREDTS} provides a conversational recommendation dataset with additional annotations for sociable recommendation strategies.
Compared with them, each dialogue in \emph{DuRecDial 2.0} attaching with seeker profiles, knowledge triples, a goal sequence is parallel in English and Chinese. 


\textbf{Multilingual and Cross-lingual Datasets for Dialog Modeling} Dialogue Systems are categorized as task-oriented and chit-chat. Several multilingual task-oriented dialogue datasets have been published \cite{mrksic-etal-2017-semantic,Schuster2019CrosslingualTL}, enabling evaluation of the approaches for cross-lingual dialogue systems. \citet{mrksic-etal-2017-semantic} annotated two languages (German and Italian) for the dialogue state tracking dataset WOZ 2.0 \cite{mrksic-etal-2017-neural} and trained a unified framework to cope with multiple languages. Meanwhile, \citet{Schuster2019CrosslingualTL} introduced a multilingual NLU dataset and highlighted the need for more sophisticated cross-lingual methods. Those datasets mainly focus on multilingual NLU and DST for task-oriented dialogue and are not parallel. In comparison with them, \emph{DuRecDial 2.0} is a bilingual parallel dataset for conversational recommendation. 
Multilingual chit-chat datasets are relatively scarce. \citet{Lin2020XPersonaEM} propose a Multilingual Persona-Chat dataset, XPersona, by extending the Persona-Chat corpora \cite{Dinan2019TheSC} to six languages: Chinese, French, Indonesian, Italian, Korean, and Japanese. In XPersona, the training sets are automatically translated using translation APIs, while the validation and test sets are annotated by human. XPersona focuses on personalized cross-lingual chit-chat generation, while \emph{DuRecDial 2.0} focuses on multilingual and cross-lingual conversational recommendation.

\section{Dataset Collection}
DuRecDial 2.0 is designed to collect highly parallel data to facilitate the study of monolingual, multilingual and cross-lingual conversational recommendation.

In this section, we describe the three steps for dataset construction: (1) Constructing the parallel data item; (2) Collecting conversation utterances by crowdsourcing; (3) Collecting knowledge triples by crowdsourcing.

\subsection{Parallel Data Item Construction}
To collect parallel data, we follow the task design in previous work \cite{liu-etal-2020-towards} and use same annotation rules, so parallel data items (e.g., knowledge graph, user profile, task templates, and conversation situation) are essential. 

\textbf{Parallel knowledge graph}
The domains covered in DuRecDial \cite{liu-etal-2020-towards} include star, movie, music, news, food, POI, and weather. As the quality of automatically translated news texts is poor, we remove the domain of news and keep other domains. For the weather domain, we construct its parallel knowledge as follows: 1) decompose Chinese weather information into some aspects of weather(e.g. the highest temperature, the lowest temperature, wind direction, etc.), 2) multiple crowdsourced annotators translate and combine English weather information to generate parallel weather information.
For other domains, the edges of knowledge graph are translated by multiple crowdsourced annotators, and the nodes are constructed 
as follows:
\begin{itemize}
    \item We crawl the English name of movies, stars, music, food, and restaurants from several related websites for the movie\footnote{https://baike.baidu.com/ \label{baike}}\footnote{http://www.mtime.com \label{mtime}}\footnote{https://maoyan.com/ \label{maoyan}}/star \textsuperscript{\ref {baike}\ref {mtime}\ref{maoyan}}/music\textsuperscript{\ref {baike}}\footnote{https://music.163.com/}\footnote{https://y.qq.com}/food\textsuperscript{\ref {baike}}\footnote{https://www.meituan.com \label{meituan}}\footnote{https://wenku.baidu.com}/POI\textsuperscript{\ref {baike}}\textsuperscript{\ref{meituan}} domain. If the English name of at least two websites is the same, it is used to construct the parallel knowledge graph.
    \item If the English names are different, crowdsourced annotators choose one of the candidate English names crawled above to construct the parallel knowledge graph.
    \item Otherwise, multiple crowdsourced annotators translate the Chinese nodes into English.
\end{itemize}

Following these rules, we finally obtain 16,556 bilingual parallel nodes and 254 parallel edges, resulting in about 123,298 parallel knowledge triplets, the accuracy of which is over 97\% \footnote{We randomly sampled 100 triplets and manually evaluated them.}. Table \ref{table:statistics} provides the statistics of \emph{DuRecDial 2.0}.

\textbf{Parallel user profiles} The user profile contains personal information (e.g. name, gender, age, residence city, occupation, etc.) and his/her preference on domains and entities. The personal information is translated by multiple crowdsourced annotators directly. The preference on domains and entities is replaced based on the parallel knowledge graph constructed above and then revised by crowdsourced annotators. 

\textbf{Parallel task templates} The task templates contain: 1) a goal sequence, where each goal consists of two elements, a dialog type and a dialog topic, corresponding to a sub-dialog, 2) a detailed description about each goal. We create parallel task templates by 1) replacing dialog type and topic based on the parallel knowledge graph constructed above, and 2) translating goal descriptions.

\textbf{Parallel conversation situation} The construction of parallel conversation situation also includes two steps: 1) decompose situation into chat time, place and topic, 
 2) multiple crowdsourced annotators translate chat time, place and topic to construct parallel conversation situation.

\subsection{Dataset Collection}

To guarantee the quality of translation, we use a strict quality control procedure. 

First, before translation, all entities in all utterances are replaced based on the parallel knowledge graph constructed above to ensure knowledge accuracy. 

Then, we randomly sample 100 conversations (about 1500 utterances) and assign them to more than 100 professional translators. After translation, all translation results are assessed 1-3 times by 3 data specialists with translation experience. Specifically, data specialists randomly select 20\% of each translator's translation results for assessment. The assessment includes word-level, utterance-level, and session-level.
For word-level assessment, they assess whether entities are consistent with the knowledge graph, whether the choice of words is appropriate, and whether there are typos. For utterance-level assessment, they assess whether the utterance is accurate, colloquial, and has no redundancy. For session-level assessment, they assess whether the session is coherent and is parallel to DuRecDial \cite{liu-etal-2020-towards}.
If the error rate exceeds 10\%, translators are no longer allowed to translate. If the error rate exceeds 3\%, we will ask translators to fix these errors. After this second-round translation, we will conduct another assessment. In second-round assessment, if the error rate is less than 2\%, translators will pass directly, otherwise, they will be assessed for the third time. In the third-round assessment, only the error rate is less than 1\% can pass. Finally, we pick 23 translators.

Finally, the 23 translators translate about 1000 utterances at a time based on the parallel user profile, knowledge graph, task templates, and conversation situation. After data translation, data specialists randomly select 10-20\% of each translator's translation results for assessment in the same way as above. The translators can continue to translate only after their passing the assessment. 

\begin{table*}[t]
	\centering
	\small
	\begin{tabular}{ p{2cm} p{7.5cm}  r } 
		\toprule[1.0pt]
		\multirow{4}{0.5in}{Knowledge graph} & \#Domains &  6\\  
		&\#Parallel entities &  16,556\\  
		&\#Parallel attributes &  254\\  
		&\#Parallel triples  &  123,298\\ 
		\hline
		\multirow{4}{0.3in}{DuRecDial 2.0} &\#Parallel dialogs&  16,482\\
		&\#Parallel sub-dialogs for QA/Rec/task/chitchat & 11,326/13,640/5,198/16,482\\
		&\#Parallel utterances &  255,346 \\
		&\#Parallel seekers &  2,714 \\
		&\#Parallel entities recommended/accepted/rejected& 17,354/13,476/3,878 \\ 
		\bottomrule[1.0pt]
		
	\end{tabular}
	\caption{Statistics of knowledge graph and \emph{DuRecDial 2.0}.}
	\label{table:statistics}
\end{table*}

\subsection{Related Knowledge Triples Annotation} Due to the complexity of this task and the massive knowledge triples corresponding to each dialog, it is very challenging for knowledge selection and goal planning. In addition to translating dialogue utterances, the annotators were also required to record the related knowledge triples if the utterances are generated according to some triples.

\section{Dataset Analysis}

\subsection{Data statistics and quality}
Table \ref{table:statistics} provides statistics of  \emph{DuRecDial 2.0} and its knowledge graph, indicating rich variability of dialog types and domains. 
Following the evaluation method in previous work \cite{liu-etal-2020-towards}, we conduct human evaluations for data quality.\footnote{A dialog will be rated ``1'' if it wholly follows the instruction in task templates and the utterances are grammatically correct and fluent, otherwise ``0''. Then we ask three persons to judge the quality of 200 randomly sampled dialogs} 
Finally we obtain an average score of 0.93 on this evaluation set. 

\subsection{Prefixes of utterances}
Since REDIAL \cite{Li2018} has been the main benchmark for conversational recommendation, we perform an in-depth comparison between the English part of \emph{DuRecDial 2.0} with REDIAL \cite{Li2018}.

\begin{figure*}[t]
		\centering
		\includegraphics[height=2.3in,width=2.3in]{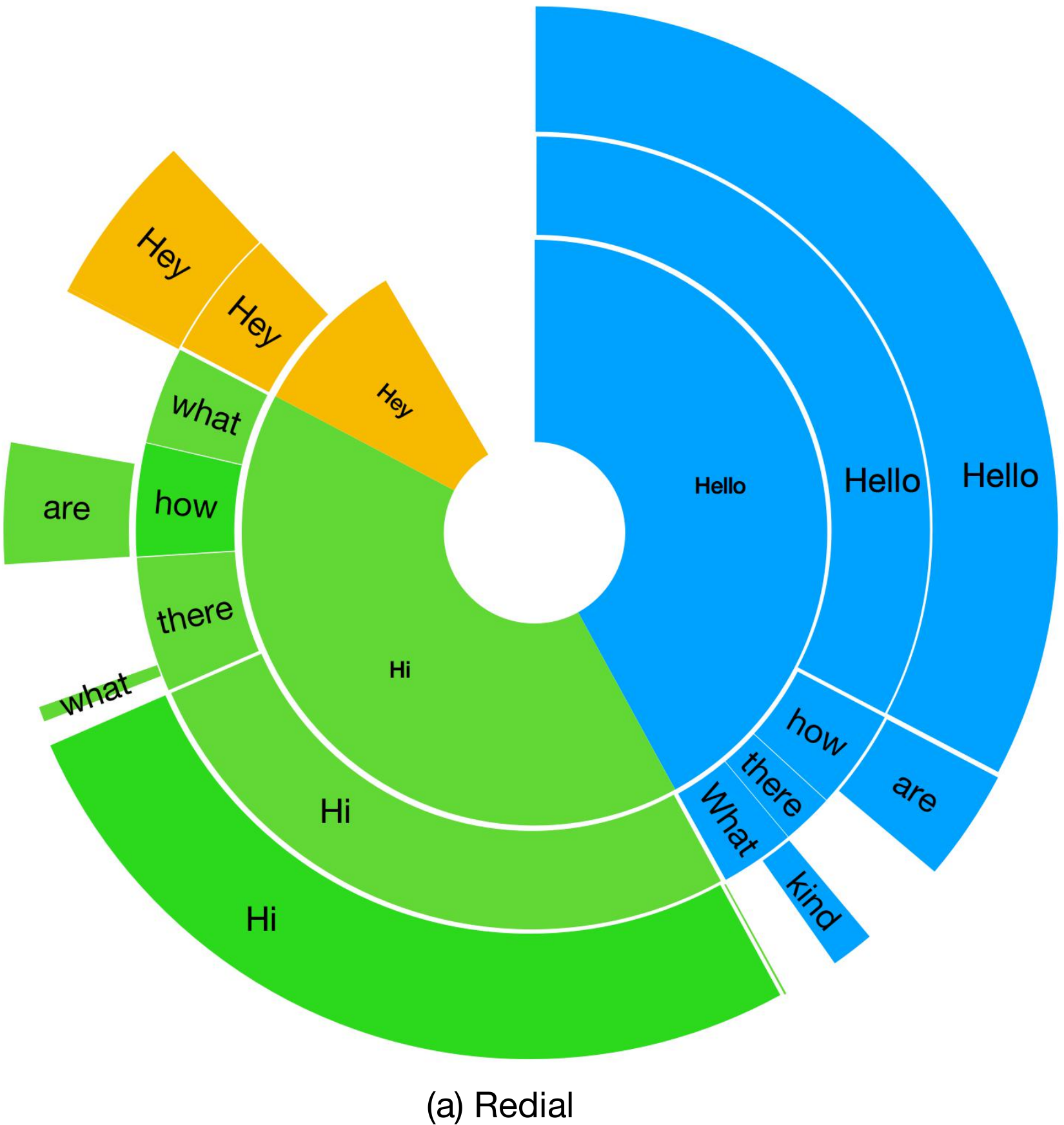}
		\includegraphics[height=2.3in,width=2.3in]{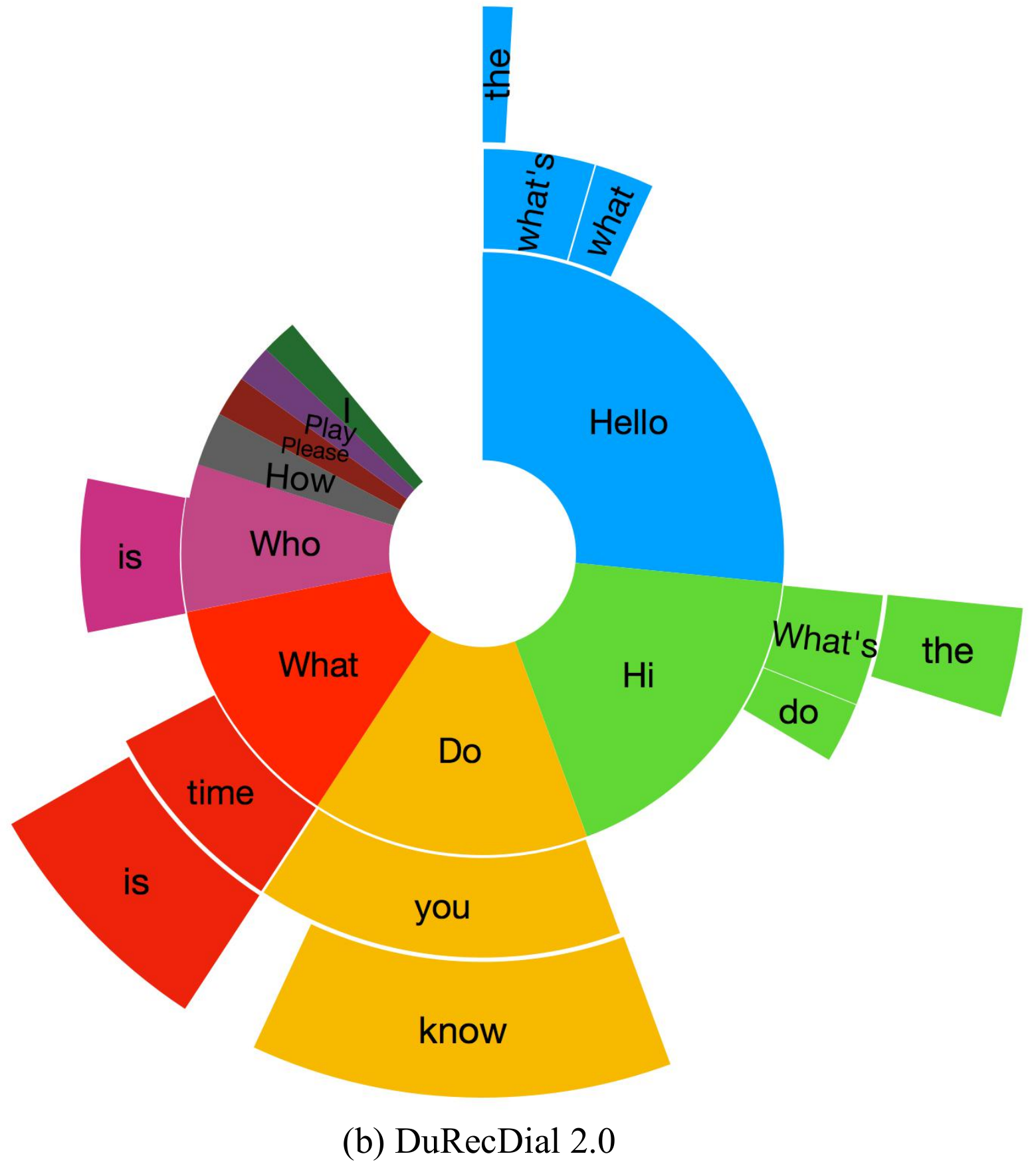}
		\caption{Distribution of trigram prefixes for first turn utterances in Redial \cite{Li2018} and \emph{DuRecDial 2.0} (Ours).}\label{fig:prefixes}
\end{figure*}

As human-bot conversations are very diversified in real-world applications, we expect a richer variability of utterances to mimic real-world application scenarios.
Figure ~\ref{fig:prefixes}(a) and Figure ~\ref{fig:prefixes}(b) show the distribution of frequent trigram prefixes.
We find that nearly all prefixes of utterances in Redial \cite{Li2018} are \emph{Hello, Hi}, and \emph{Hey}, while the prefixes of utterances in \emph{DuRecDial 2.0} are more diversified. For example, several sectors indicated by prefixes \emph{Do, What, Who, How, Please, Play}, and \emph{I} are frequent in \emph{DuRecDial 2.0} but are completely absent in Redial \cite{Li2018}, indicating that \emph{DuRecDial 2.0} has a more flexible language style.

\section{Task Formulation on \emph{DuRecDial 2.0}}

Let $D^k=\{d^k_{i}\}^{N_{D^k}}_{i=0}$ denote a set of dialogs by the seeker $s_k$ $(0\le k<N_s)$, where $N_{D^k}$ is the number of dialogs by the seeker $s_k$, and $N_s$ is the number of seekers. 
Recall that we attach each dialog (say $d^k_{i}$) with an updated seeker profile (denoted as $\mathcal{P}^{s_k}_{i}$), a knowledge graph $\mathcal{K}=\{k_j\}_{j=0}^{m}$, a goal sequence $\mathcal{G}=\{(g^{ty}_{j}, g^{tp}_{j})\}^{m}_{j=0}$, where $k_j$ is several knowledge triples, $g^{ty}_{j}$ is a candidate dialog type and $g^{tp}_{j}$ is a candidate dialog topic.
Given a context $X$ with utterances $\{u_j\}_{j=0}^{m-1}$ from the dialog $d^k_{i}$,  $\mathcal{G}$, $\mathcal{P}^{s_k}_{i}$ and $\mathcal{K}$, the aim is to produce a proper response $Y=u_m$ for completion of the goal $g_c=(g^{ty}_{m}, g^{tp}_{m})$. 

\paragraph{Monolingual conversational recommendation:} \textbf{Task 1:} ($X_{en}$, $\mathcal{G}_{en}$, $\mathcal{K}_{en}$, $Y_{en}$) or \textbf{Task 2:} ($X_{zh}$, $\mathcal{G}_{zh}$, $\mathcal{K}_{zh}$, $Y_{zh}$).
With these two monolingual conversational recommendation forms, we can investigate the performance variation of the same model trained on two separate datasets in different languages. In our experiments, we train two conversational recommendation models respectively for the two monolingual tasks. Then we can evaluate their performance variation across English and Chinese to see how the changes between languages can affect model performance.

\paragraph{Multilingual conversational recommendation:} \textbf{Task 3:} ($X_{en}$, $\mathcal{G}_{en}$, $\mathcal{K}_{en}$, $Y_{en}$, $X_{zh}$, $\mathcal{G}_{zh}$, $\mathcal{K}_{zh}$, $Y_{zh}$).
Similar to multilingual neural machine translation\cite{johnson-etal-2017-googles} and multilingual reading comprehension\cite{jing-etal-2019-bipar},  we directly mix training instances of the two languages into a single training set and train a single model to handle both English and Chinese conversational recommendation at the same time. This task setting can help us investigate if the use of additional training data in another language can bring performance benefits for a model of current language.  

\paragraph{Cross-lingual conversational recommendation:} 
The two forms of crosslingual conversational recommendation are \textbf{Task 4:} ($X_{zh}$, $\mathcal{G}_{en}$, $\mathcal{K}_{en}$, $Y_{en}$) and \textbf{Task 5:} ($X_{en}$, $\mathcal{G}_{zh}$, $\mathcal{K}_{zh}$, $Y_{zh}$), where given related goals and knowledge (e.g., in Engish), the model takes dialog context in one language (e.g., in Chinese) as input, and then produce responses in another language (e.g., in Engish) as output. Understanding the mixed-language dialog context is a desirable skill for end-to-end dialog systems. This task setting can help evaluate if a model has the capability to perform this kind of cross-lingual tasks. 

\section{Experiments and Results}

    \subsection{Experiment Setting}
    \textbf{Dataset} For the train/development/test set, we follow the split of  \cite{liu-etal-2020-towards}, with one notable difference that we discard the dialogues that include news.
    
    \textbf{Automatic Evaluation Metrics:} For automatic evaluation from the viewpoint of conversation, we follow the setting in previous work \cite{liu-etal-2020-towards} to use several common metrics such as F1, BLEU (DLEU1 and DLEU2) \cite{Papineni2002}, and DISTINCT (DIST-1 and DIST-2) \cite{li2016diversity} to measure the relevance, fluency, and diversity of generated responses. Moreover, we also evaluate the knowledge-selection capability of each model by calculating knowledge precision/recall/F1 scores as done in \citet{Wu2019, liu-etal-2020-towards}.\footnote{When calculating the knowledge precision/recall/F1, we compare the generated results with the correct knowledge.}
    In addition, to evaluate recommendation effectiveness, we design two automatic metrics shown as follows. First, to measure how well a model can lead the whole dialog to approach a recommendation target, we design a metric \emph{dialog-Leading Success rate}~(\textbf{\emph{LS }}). It calculates the percentage of times a dialog can successfully reach or mention the target after a few dialog turns.\footnote{We convert each multi-turn dialog into multiple (context, response) pairs, and generate a response for each context, and then evaluate LS and UTC based on the generated conversation (dialog level).} Second, to measure how well a model can respond to new topics by users, we design a metric \emph{User-Topic Consistency rate}~(\textbf{\emph{UTC}}). It calculates the percentage of times the model can successfully follow new topics mentioned by users.\footnote{If the generated response is coherent with the new topic mentioned by user, we define "successfully follow the topic".}
    
    \textbf{Human Evaluation Metrics:} The human evaluation is conducted at the level of both turns and dialogs. 
	
	For turn-level human evaluation, we ask each model to produce a response conditioned on a given context, goal and related knowledge. The generated responses are evaluated by three persons in terms of fluency, appropriateness, informativeness, proactivity, and knowledge accuracy.\footnote{Please see supplemental material for more details.} 
	
	 For dialogue-level human evaluation, we let each model converse with humans and proactively make recommendations when given goals and reference knowledge. For each model, we collect 30 dialogs. These dialogs are then evaluated by three persons in terms of two metrics: (1) coherence that examines fluency, relevancy and logical consistency of each response when given the current goal and context, and (2) recommendation success rate that measures measures the percentage of times users finally accept the recommendation at the end of a dialog.
	
	The evaluators rate the dialogs on a scale of 0 (poor) to 2 (good) in terms of each human metric except \textbf{\emph{recommendation success rate}}.\footnote{Please see supplemental material for more details.}

\begin{table*}
		\centering
		\small
		\begin{tabular}{c c c c c c c c}
			\toprule[1.0pt]
			Tasks & Methods  &  F1  &BLEU1/ BLEU2 &  DIST-1/DIST-2 & Knowledge P/R/F1   & LS  & UTC \\
			\toprule[1.0pt]
			1(EN->EN)& XNLG   & 43.78\%  & 0.202/ 0.123   &   0.016/ 0.053 &   0.173/0.211/0.179 &15.19\% &11.44\%\\
			2(ZH->ZH)& XNLG   & 36.58\%  & 0.319/ 0.213   &   0.010/0.033 &   0.327/0.401/0.352 &22.17\% &20.29\% \\ 
			3(EN->EN)& XNLG   & 42.03\%  & 0.199/ 0.131   &   0.008/ 0.021 &   0.171/0.207/0.173 &10.03\% &11.31\% \\
			3(ZH->ZH)& XNLG   & 36.61\%  & 0.322/ 0.208  &   0.006/ 0.020 &   0.324/0.393/0.351 &22.03\% &20.31\% \\
			4(ZH->EN)& XNLG   & 41.98\%  & 0.201/ 0.129  &   0.019/ 0.075 &   0.123/0.162/0.139 &14.81\% &13.61\%\\
			5(EN->ZH)& XNLG   & 36.53\%  & 0.323/ 0.202   &   0.014/ 0.052 &  0.308/0.394/0.318 &21.43\% &20.06\%\\ \hline
			1(EN->EN)& mBART   & \textbf{66.96\%}  & 0.285/ 0.195  &   0.018/ 0.057 &   0.276/0.313/0.285 &21.03\% &20.26\%\\ 
			2(ZH->ZH)& mBART   & 46.29\%  & 0.363/ 0.259  &   0.011/0.042 &   0.416/0.491/0.432 &35.89\% &31.98\% \\   
			3(EN->EN)& mBART   & 63.69\%  & 0.254/ 0.168   &   0.008/ 0.023 &   0.253/0.300/0.266 &13.11\% &18.55\% \\ 
			3(ZH->ZH)& mBART   & 46.23\%  & 0.368/ 0.237  &   0.006/ 0.024 &   \textbf{0.432}/0.499/\textbf{0.451} & 35.81\% & \textbf{31.99\%} \\ 
			4(ZH->EN)& mBART   & 64.31\%  & 0.267/ 0.185   &   \textbf{0.027}/ 0.084 &   0.229/0.261/0.236 &21.37\% &22.79\% \\ 
			5(EN->ZH)&mBART   & 53.55\%  &  \textbf{0.392} / \textbf{0.304}  &   0.026/ \textbf{0.097} &  0.421/\textbf{0.514}/0.439 & \textbf{37.04\%}  &30.86\% \\ 
			\bottomrule[1.0pt]
			
		\end{tabular}
		\caption{Automatic evaluation results on parallel corpus of automatic translation. Task 1-5 represent the 5 different tasks on \emph{DuRecDial 2.0}: ($X_{en}$, $\mathcal{G}_{en}$, $\mathcal{K}_{en}$, $Y_{en}$), ($X_{zh}$, $\mathcal{G}_{zh}$, $\mathcal{K}_{zh}$, $Y_{zh}$),  ($X_{en}$, $\mathcal{G}_{en}$, $\mathcal{K}_{en}$, $Y_{en}$, $X_{zh}$, $\mathcal{G}_{zh}$, $\mathcal{K}_{zh}$, $Y_{zh}$), ($X_{zh}$, $\mathcal{G}_{en}$, $\mathcal{K}_{en}$, $Y_{en}$), and ($X_{en}$, $\mathcal{G}_{zh}$, $\mathcal{K}_{zh}$, $Y_{zh}$). Task 1 and 2 are monolingual, task 3 is multilingual, and task 4 and 5 are cross-lingual.``EN'',  and ``ZH'' stands for English, and Chinese respectively.}
		\label{table:auto-results1}
	\end{table*}

\begin{table*}
		\centering
		\small
		\begin{tabular}{c c c c c c c c}
			\toprule[1.0pt]
			Tasks & Methods  &  F1  &BLEU1/ BLEU2 &  DIST-1/DIST-2 & Knowledge P/R/F1   & LS  & UTC \\  
			\toprule[1.0pt]
			1(EN->EN)& XNLG   & 49.66\%  & 0.265/ 0.173   &   0.018/ 0.050 &   0.244/0.291/0.260 &16.31\%  &15.18\% \\ 
			2(ZH->ZH)& XNLG   & 36.58\%  & 0.319/ 0.213   &   0.010/0.033 &   0.327/0.401/0.352 &22.17\% &20.29\% \\ 
			3(EN->EN)& XNLG   & 44.15\%  & 0.202/ 0.142   &   0.009/ 0.021 &   0.173/0.211/0.185 &13.01\% &12.31\% \\
			3(ZH->ZH)& XNLG   & 36.62\%  & 0.329/ 0.182  &   0.008/ 0.023 &   0.328/0.405/0.359 &22.29\% &20.62\% \\
			4(ZH->EN)& XNLG   & 45.75\%  & 0.239 / 0.171 &   0.013/ 0.036 &  0.217/0.259/0.211 &14.55\% &15.03\% \\
			5(EN->ZH)& XNLG   & 36.77\%  & 0.330/ 0.203  &   0.011/ 0.053 &  0.331/0.393/0.355  &21.57\% &20.17\% \\\hline
			1(EN->EN)& mBART   & \textbf{68.38\%}  & 0.325/ 0.245  &   0.017/ 0.054 &   0.350/0.396/0.362 &28.90\%  &24.77\% \\ 
			2(ZH->ZH)& mBART   & 46.29\%  & 0.363/ 0.259  &   0.011/0.042 &   0.416/0.491/0.432 &35.89\% &31.98\% \\  
			3(EN->EN)& mBART   & 64.38\%  & 0.268/ 0.192  &   0.007/ 0.024 &   0.307/0.367/0.325 &16.11\% &20.55\% \\ 
			3(ZH->ZH)& mBART   & 46.37\%  & 0.366/ 0.241  &   0.006/ 0.025 &   0.412/0.493/0.436 &36.31\% & \textbf{32.59\%} \\ 
			4(ZH->EN)& mBART   & 67.43\%  & 0.314/ 0.231  &   0.013/ 0.040 &  0.328/0.379/0.343   &24.26\% &23.83\% \\ 
			5(EN->ZH)& mBART   & 55.69\%  & \textbf{0.430} /  \textbf{0.325}   &   \textbf{0.019}/ \textbf{0.077} &  \textbf{0.455}/\textbf{0.536}/\textbf{0.476}   & \textbf{38.67\%} &32.11\% \\ 
			\bottomrule[1.0pt]
			
		\end{tabular}
		\label{table:auto-results}
		\caption{Automatic evaluation results on \emph{DuRecDial 2.0}. Task 1-5 represent the 5 different tasks on \emph{DuRecDial 2.0}: ($X_{en}$, $\mathcal{G}_{en}$, $\mathcal{K}_{en}$, $Y_{en}$), ($X_{zh}$, $\mathcal{G}_{zh}$, $\mathcal{K}_{zh}$, $Y_{zh}$),  ($X_{en}$, $\mathcal{G}_{en}$, $\mathcal{K}_{en}$, $Y_{en}$, $X_{zh}$, $\mathcal{G}_{zh}$, $\mathcal{K}_{zh}$, $Y_{zh}$), ($X_{zh}$, $\mathcal{G}_{en}$, $\mathcal{K}_{en}$, $Y_{en}$), and ($X_{en}$, $\mathcal{G}_{zh}$, $\mathcal{K}_{zh}$, $Y_{zh}$). Task 1 and 2 are monolingual, task 3 is multilingual, and task 4 and 5 are cross-lingual. ``EN'',  and ``ZH'' stands for English, and Chinese respectively.}
		\label{table:auto-results}
	\end{table*}

\begin{table*}
		\centering
		\small
		\begin{tabular}{ c c r r r r  r c c} 
			\toprule[1.0pt]
			 & \multicolumn{6}{c|}{Turn-level results}  & \multicolumn{2}{c}{Dialog-level results} \\ \cmidrule{3-7} \cmidrule{8-9}
			Tasks & Methods & Fluency  & Appro. & Infor.  & Proactivity & \multicolumn{1}{c|}{Know. Acc.} & Coherence  & Rec. success rate \\ 
			\toprule[1.0pt] 
			1(EN->EN)& XNLG  & 1.96 & 1.09 & 0.33 & 1.08 & 0.68 &0.31 & 13\%  \\
			2(ZH->ZH)& XNLG  & 1.94 & 1.16 & 0.37 & 1.03 & 0.68 &0.38 & 23\%  \\
			3(EN->EN)& XNLG  & 1.95 & 0.98 & 0.29 & 0.97 & 0.42 &0.22 & 10\%  \\
			3(ZH->ZH)& XNLG  & 1.93 & 1.18 & 0.39 & 1.01 & 0.61 &0.40 & 27\%  \\
			4(ZH->EN)& XNLG  & 1.94 & 1.07 & 0.34 & 1.02 & 0.44 &0.29 & 10\%  \\
			5(EN->ZH)& XNLG  & 1.95 & 1.15 & 0.41 & 1.05 & 0.62 &0.42 & 27\%  \\ \hline
			1(EN->EN)& mBART  & 1.97 & 1.21 & 0.46 & 1.19 & 0.68 &0.46 & 17\%  \\
			2(ZH->ZH)& mBART  & 1.97 & 1.22 & 0.51 & 1.15 & 0.68 &0.52 & 20\%  \\
			3(EN->EN)& mBART  & 1.97 & 1.06 & 0.44 & 1.01 & 0.51 &0.37 & 10\%  \\
			3(ZH->ZH)& mBART  & \textbf{1.98} & 1.27 & \textbf{0.53} & 1.18 & 0.55 &\textbf{0.68} & 23\%  \\
			4(ZH->EN)& mBART  & 1.96 & 1.17 & 0.46 & 1.10 & 0.53 &0.41 & 13\% \\
			5(EN->ZH)& mBART  & 1.97 & \textbf{1.29} & 0.52 & \textbf{1.21} & \textbf{0.79} &0.60 & \textbf{33\%}  \\
			\bottomrule[1.0pt]
		\end{tabular}
		\caption{Human evaluation results  on \emph{DuRecDial 2.0} at the level of turns and dialogs.  ``Appro.'',  ``Infor.'',  ``Know. Acc.'',  ``Rec.'', ``EN'',  and ``ZH'' stands for appropriateness, informativeness, knowledge accuracy, recommendation, English, and Chinese respectively. Task 1-5 represent the 5 different tasks on \emph{DuRecDial 2.0}: ($X_{en}$, $\mathcal{G}_{en}$, $\mathcal{K}_{en}$, $Y_{en}$), ($X_{zh}$, $\mathcal{G}_{zh}$, $\mathcal{K}_{zh}$, $Y_{zh}$),  ($X_{en}$, $\mathcal{G}_{en}$, $\mathcal{K}_{en}$, $Y_{en}$, $X_{zh}$, $\mathcal{G}_{zh}$, $\mathcal{K}_{zh}$, $Y_{zh}$), ($X_{zh}$, $\mathcal{G}_{en}$, $\mathcal{K}_{en}$, $Y_{en}$), and ($X_{en}$, $\mathcal{G}_{zh}$, $\mathcal{K}_{zh}$, $Y_{zh}$). Task 1 and 2 are monolingual, task 3 is multilingual, and task 4 and 5 are cross-lingual.}
		\label{table:human-results}
	\end{table*}

	\subsection{Methods}
	
	\textbf{XNLG} \cite{Chi2020CrossLingualNL} 
	is a cross-lingual pre-trained model with both monolingual and cross-lingual objectives and updates the parameters of the encoder and decoder through auto-encoding and autoregressive tasks to transfer monolingual NLG supervision to other pre-trained languages.
	When the target language is the same as the language of training data, we fine-tune the parameters of encoder and decoder. When the target language is different from the language of training data, we fine-tune the the parameters of encoder. The objective of fine-tuning encoder is to minimize:

	\begin{equation}\nonumber
	\begin{aligned}
	\mathcal{L}_e = \sum_{(x, y) \in \mathcal{D}_p} \mathcal{L}^{(x,y)}_{XMLM}  + \sum_{(x) \in \mathcal{D}_m} \mathcal{L}^{(x)}_{MLM}
    \end{aligned}
	\end{equation}
	where $\mathcal{L}^{(x,y)}_{XMLM}$  and $\mathcal{L}^{(x)}_{MLM}$ are the same as XNLG, $\mathcal{D}_p$ indicates the parallel corpus, and  $\mathcal{D}_m$ is the monolingual corpus.
	
	The objective of fine-tuning decoder is to minimize:
	\begin{equation}\nonumber
	\begin{aligned}
	\mathcal{L}_d = \sum_{(x, y) \in \mathcal{D}_p} \mathcal{L}^{(x,y)}_{XAE}  + \sum_{(x) \in \mathcal{D}_m} \mathcal{L}^{(x)}_{DAE}
    \end{aligned}
	\end{equation}
	where $\mathcal{L}^{(x,y)}_{XAE}$  and $\mathcal{L}^{(x)}_{EAE}$ are the same as XNLG.


    \textbf{mBART} \cite{Liu2020MultilingualDP} is a multilingual sequence-to-sequence (Seq2Seq) denoising auto-encoder pre-trained on a subset of 25 languages – CC25 – extracted from the Common Crawl (CC) \cite{wenzek-etal-2020-ccnet,conneau-etal-2020-unsupervised}. It provides a set of parameters that can be fine-tuned for any of the language pairs in CC25 including English and Chinese. Loading mBART initialization can provide performance gains for monolingual/multilingual/cross-lingual tasks and serves as a strong baseline.

    We treat our 5 tasks as Machine Translation(MT) task. Specifically, context, knowledge, and goals are concatenated as source language input, which could be monolingual, multilingual, or cross-lingual text, then the corresponding response is generated as the target language output. 
    Since the response could be in different languages, we also concatenate a language identifier of response to the source input. 
    Concretely,  if the response is in English, the identifier is \emph{EN}, otherwise \emph{ZH}, no matter what language the source input is. 
    We finally fine-tune the mBART model on our 5 tasks respectively.

	\subsection{Experiment Results}
	Table \ref{table:auto-results1} and Table \ref{table:auto-results} presents automatic evaluation results on automatic translation \footnote{We use https://fanyi.baidu.com/} parallel corpus and human translation parallel corpus (\emph{DuRecDial 2.0}). Table \ref{table:human-results} provides human evaluation results on \emph{DuRecDial 2.0}.
	
	\textbf{Automatic Translation vs. Human Translation:} As shown in Table \ref{table:auto-results1} and Table \ref{table:auto-results}, the models of XNLG \cite{Chi2020CrossLingualNL} and mBART \cite{Liu2020MultilingualDP} trained with human-translated parallel corpus (DuRecDial 2.0) are both better than those trained with machine-translated parallel corpus across almost all the tasks. 
	The possible reason is that automatic translation might contain many translation errors, which increases the difficulty for effective learning by models.
	
	\textbf{English vs. Chinese:} As shown in Table \ref{table:auto-results} and \ref{table:human-results}, the results of Chinese related tasks (Task 2, 3(ZH->ZH), 5) are better than that for English related tasks (Task 1, 3(EN->EN), 4) in terms of almost all the metrics, except for \emph{F1} and \emph{DIST1/DIST2}. 
	The possible reason is that: (1) most of entities in this dataset are from the domain of Chinese movies and famous Chinese entertainers, which are quite different from the set of entities in English pretraining corpora used for XNLG or mBART; (2) then the pretrained models perform poorly for the modeling of these entities in utterances, resulting in knowledge errors in responses (e.g., the agent might mention incorrect entities in responses that are not relevant to current topic), since some entities might never appear in the English pretraining corpora. The accuracy of generated entities in responses is very crucial to model performance in terms of \emph{Knowledge P/R/F1}, \emph{LS}, \emph{UTC}, \emph{Know. Acc.}, \emph{Coherence}, and \emph{Rec. success rate}. Therefore incorrect entities in generated responses deteriorate model performance in terms of the above metrics for English related tasks.
	
	\textbf{Monolingual vs. Multilingual:} 
	Based on the results in Table \ref{table:auto-results} and \ref{table:human-results}, the model for multilingual Chinese task (Task 3(ZH->ZH)) are better than the monolingual Chinese model (Task 2) in terms of almost all the metrics (except for \emph{DISTINCT} and \emph{Knowledge Accuracy}). It indicates that the use of additional English corpora can slightly improve model performance for Chinese conversational recommendation. The possible reason is that the use of additional English data implicitly expands the training data size for Chinese related tasks through the bilingual training paradigm of XNLG or mBART, which strengthens the capability of generating correct entities for a given dialog context. Then Chinese related task models can generate correct entities in responses more frequently, leading to better model performance.
	
	But the model for multilingual English task (Task 3(EN->EN)) can not outperform the monolingual English model (Task 1). The possible reason is that the pretrained models can not perform well on the modeling of entities in dialog utterances, resulting in poor model performance.
	
	\textbf{Monolingual vs. Cross-lingual:} 
	According to the results in Table \ref{table:auto-results} and \ref{table:human-results}, the model of EN->ZH cross-lingual task (Task 5) perform surprisingly better than the monolingual Chinese model (Task 2) in terms of all the automatic and human metrics (except for \emph{Fluency}) (sign test, p-value \textless 0.05). It indicates that the use of bilingual corpora can consistently bring performance improvement for Chinese conversational recommendation. One possible reason is that XNLG or mBART can fully exploit the bilingual dataset, which strengthens the capability of generating correct entities in responses for Chinese related tasks. Moreover, we notice that the model performance is further improved from the multilingual setting to the cross-lingual setting, and the reason for this result will be investigated in the future work. 
	
	But the ZH->EN cross-lingual model (Task 4) can not outperform the monolingual English model (Task 1), which is consistent with the results with the multilingual setting. 

	
    
	\textbf{XNLG vs. mBART:} According to the evaluation results in Table \ref{table:auto-results1}, Table \ref{table:auto-results} and Table \ref{table:human-results}, mBART \cite{Liu2020MultilingualDP} outperforms XNLG \cite{Chi2020CrossLingualNL} across almost all the tasks or metrics. The main reason is that mBART employs more model parameters and it uses more parallel corpora for training when compared with XNLG.
	
	

\section{Conclusion}
To facilitate the study of multilingual and cross-lingual conversational recommendation, we create a bilingual parallel dataset \emph{DuRecDial 2.0} and define 5 tasks on it. We further establish baselines for monolingual, multilingual, and cross-lingual conversational recommendation. 
Automatic evaluation and human evaluation results show that our bilingual dataset, \emph{DuRecDial 2.0}, can bring performance improvement for Chinese conversational recommendation. 
Besides, \emph{DuRecDial 2.0} provides a challenging testbed for future studies of monolingual, multilingual, and cross-lingual conversational recommendation.
In future work, we will investigate the possibility of combining multilinguality and few (or zero) shot learning to see if it can help dialog tasks in low-resource languages.

\section{Ethical Considerations}
We make sure that \emph{DuRecDial 2.0}  was collected in a manner that is consistent with the terms of use of any sources and the intellectual property and privacy rights of the original authors of the texts. 
And crowd workers were treated fairly. This includes, but is not limited to, compensating them fairly, ensuring that they were able to give informed consent, and ensuring that they were voluntary participants who were aware of any risks of harm associated with their participation.
Please see Section 3 and 4 for more details characteristics and collection process of \emph{DuRecDial 2.0}.

\section*{Acknowledgments}

Thanks for the insightful comments from reviewers and the support of dataset construction from Ying Chen. This work is supported by the National Key Research and Development Project of China (No.2018AAA0101900) and the Natural Science Foundation of China (No. 61976072).

\bibliography{emnlp2021}
\bibliographystyle{emnlp2021}

\clearpage
\section*{Appendix}

\subsection*{1. Turn-level Human Evaluation Guideline}
\textbf{Fluency} measures fluency of each response: 
\begin{itemize}
\item score 0 (bad): unfluent and difficult to understand.
\item score 1 (fair): there are some errors in the response text but still can be understood.
\item score 2 (good): fluent and easy to understand. 
\end{itemize}

\noindent
\textbf{Appropriateness} examines relevancy of each response when given the current goal and local context:
\begin{itemize}
\item score 0 (bad): not relevant to the current goal and context.
\item score 1 (fair): relevant to the current goal and context, but using some irrelevant knowledge.
\item score 2 (good): otherwise.
\end{itemize}

\noindent

\noindent
\textbf{Informativeness} examines how much knowledge (goal topics and topic attributes) is provided in responses:
 \begin{itemize}
 \item score 0 (bad): no knowledge is mentioned at all.
 \item score 1 (fair): only one knowledge triple is mentioned in the response.
 \item score 2 (good): more than one knowledge triple is mentioned in the response.\\
\end{itemize}

\noindent

\textbf{Proactivity} measures how well the model can introduce new topics with good fluency and relevance:
\begin{itemize}
\item score 0 (bad): some new topics are introduced but irrelevant to the context.
\item score 1 (fair): no new topics/knowledge are used.
\item score 2 (good): some new topics relevant to the context are introduced.
\end{itemize}

\noindent
\textbf{Knowledge accuracy} evaluates correctness of the knowledge in responses:
\begin{itemize}
\item score 0 (bad): all knowledge used is wrong, or no knowledge is used.
\item score 1 (fair): part of the knowledge used is correct.
\item score 2(good): all knowledge used is correct.\\
\end{itemize}

\noindent

\subsection*{2. Dialogue-level Human Evaluation Guideline}

\textbf{Coherence} measures fluency, relevancy and logical consistency of each response when given the current goal and global context:
\begin{itemize}
\item score 0 (bad): more than two-thirds responses irrelevant or logical contradictory to the given current goal and global context.
\item score 1 (fair): more than one-third responses irrelevant or logical contradictory to the given current goal and global context.
\item score 2 (good): otherwise.
\end{itemize}

\noindent
\textbf{Recommendation success rate} measures the percentage of times users finally accept the recommendation at the end of a dialog:
\begin{itemize}
\item score 0 (bad): user not accept the recommendation.
\item score 1 (good): user finally accept the recommendation.
\end{itemize}

\noindent

\subsection*{3. Case Study}
\begin{figure*}[h]
        \centering
        \subfigure[Case generated by mBART for task 1-3.]{
	    \includegraphics[height=4.5in,width=6.1in]{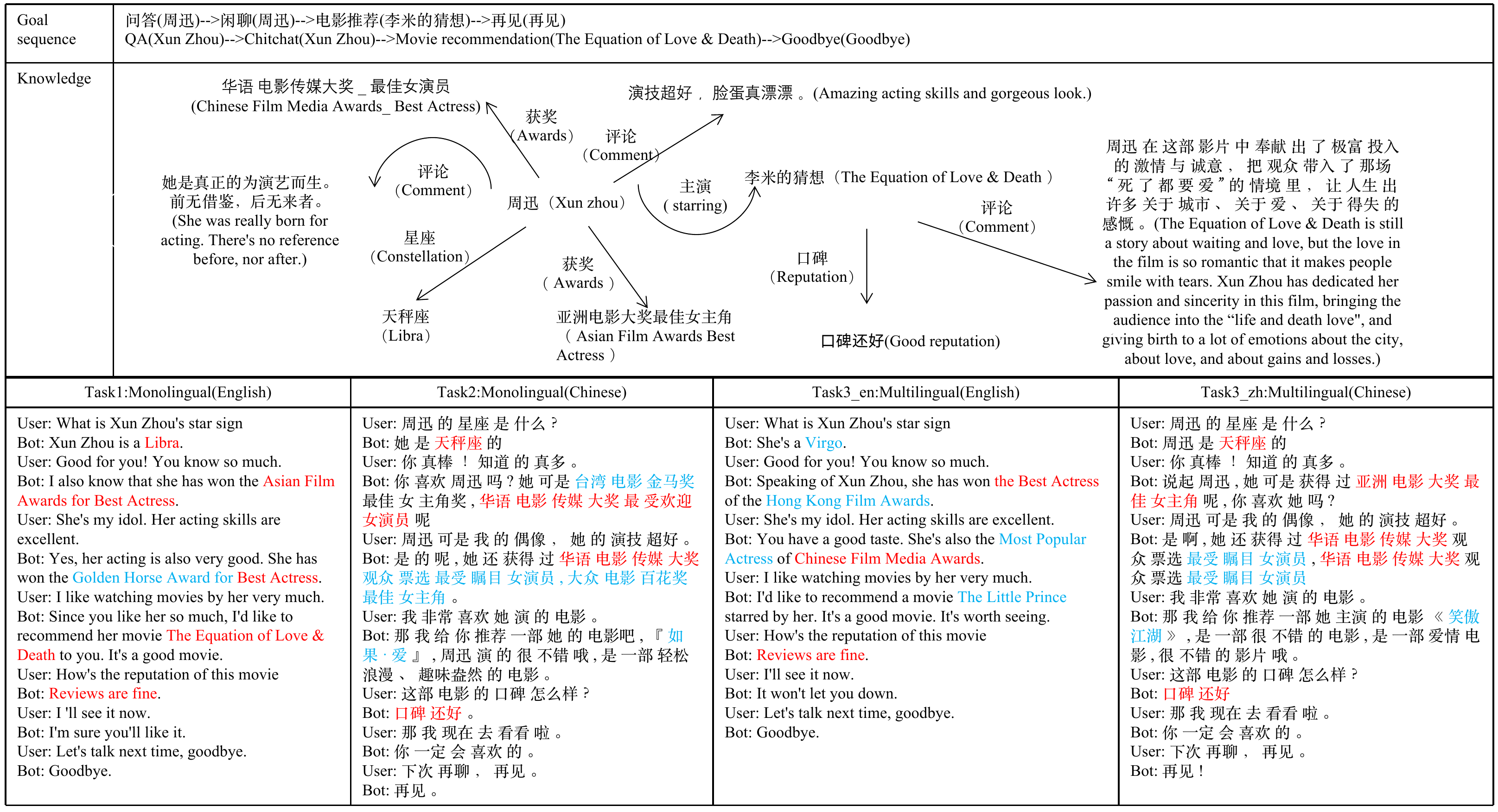}}
	    \subfigure[Case generated by mBART for task 4-5.]{
	    \includegraphics[height=4.5in,width=6.1in]{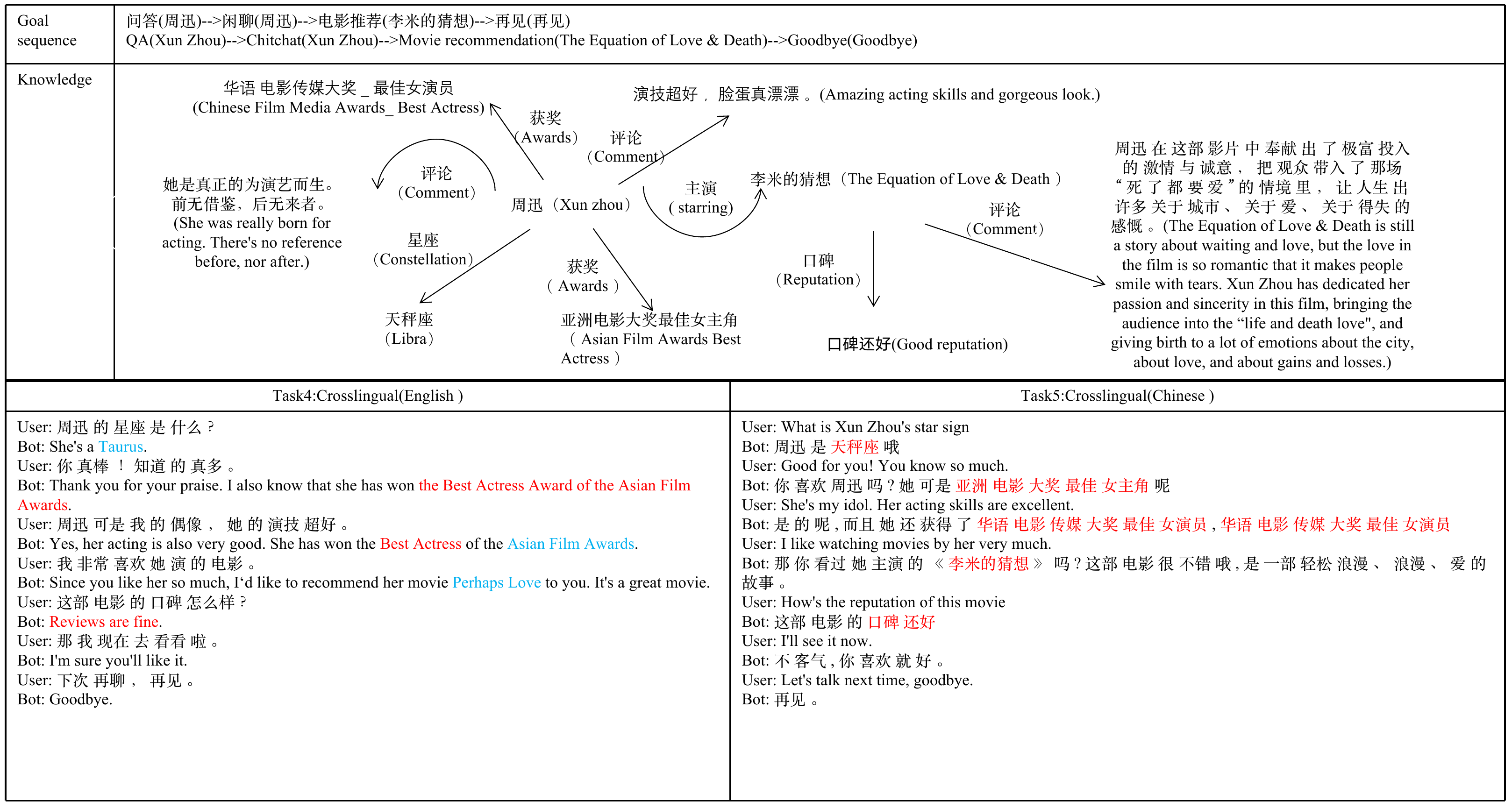}}
        \caption{Conversations generated by mBART: texts in red color represent correct knowledge being appropriate in current context, while texts in blue color represent inappropriate knowledge. }
        \label{fig:case}
\end{figure*}

Figure ~\ref{fig:case} shows the conversations generated by mBART via conversing with humans, given the conversation goal and the related knowledge.  It can be seen that the use of additional English data can bring performance improvement for Chinese conversational recommendation, especially in terms of \emph{Knowledge P/R/F1}.

\end{document}